%% file: main.tex
\pdfoutput=1
\documentclass[10pt,twocolumn,letterpaper]{article}

\usepackage[hyperref]{acl2020}
\usepackage{times}
\usepackage{epsfig}
\usepackage{graphicx}
\usepackage{amsmath}
\usepackage{amssymb}

\usepackage{multirow}
\usepackage{booktabs}
\usepackage{xcolor}
\usepackage{listings}
\usepackage{soul}
\usepackage{cleveref}

\usepackage{bm}
\usepackage{xspace}
\input{math_commands.tex}

\aclfinalcopy

\title{A Multi-modal Approach to Fine-grained Opinion Mining \\ on Video Reviews}

\author{
Edison Marrese-Taylor\textsuperscript{1}\textsuperscript{*}, Cristian Rodriguez-Opazo\textsuperscript{2}\textsuperscript{*}, Jorge A. Balazs\textsuperscript{1}
  \\ {\bf Stephen Gould\textsuperscript{2}} and {\bf Yutaka Matsuo\textsuperscript{1}} \\
Graduate School of Engineering, The University of Tokyo, Japan\textsuperscript{1} \\
{\tt \{emarrese, jorge, matsuo\}@weblab.t.u-tokyo.ac.jp}\\
Australian Centre for Robotic Vision (ACRV), Australian National University\textsuperscript{2}\\ 
{\tt \{cristian.rodriguez, stephen.gould\}@anu.edu.au}\\
\footnotesize{\textsuperscript{*}Authors contributed equally to this work.}}

\date{}
\begin{document}
\maketitle


\begin{abstract}
    Despite the recent advances in opinion mining for written reviews, few works have tackled the problem on other sources of reviews. In light of this issue, we propose a multi-modal approach for mining fine-grained opinions from video reviews that is able to determine the aspects of the item under review that are being discussed and the sentiment orientation towards them.
    Our approach works at the sentence level without the need for time annotations and uses features derived from the audio, video and language transcriptions of its contents.
    We evaluate our approach on two datasets and show that leveraging the video and audio modalities consistently provides increased performance over text-only baselines, providing evidence these extra modalities are key in better understanding video reviews.

\end{abstract}

\section{Introduction}

Sentiment analysis (SA) is an important task in natural language processing, aiming at identifying and extracting opinions, emotions, and subjectivity. As a result, sentiment can be automatically collected, analyzed and summarized. Because of this, SA has received much attention not only in academia but also in industry, helping provide feedback based on customers' opinions about products or services. The underlying assumption in SA is that the entire input has an overall polarity, however, this is usually not the case. For example, laptop reviews generally not only express the overall sentiment about a specific model (e.g., ``This is a great laptop''), but also relate to its specific aspects, such as the hardware, software or price. Subsequently, a review may convey opposing sentiments (e.g., ``Its performance is ideal, I wish I could say the same about the price'') or objective information (e.g., ``This one still has the CD slot'') for different aspects of an entity. Aspect-based sentiment analysis (ABSA) or fine-grained opinion mining aims to extract opinion targets or aspects of entities being reviewed in a text, and to determine the sentiment reviewers express for each. 
ABSA allows us to evaluate aggregated sentiments for each aspect of a given product or service and gain a more granular understanding of their quality. This is of especial interest for companies as it enables them to refine specifications for a given product or service, and leading to an improved  overall customer satisfaction.

Fine-grained opinion mining is also important for a variety of NLP tasks, including opinion-oriented question answering and opinion summarization. In practical terms, the ABSA task can be divided into two sub-steps, namely aspect extraction (\textit{AE}) and (aspect level) sentiment classification (\textit{SC}), which can be tackled in a pipeline fashion, or simultaneously (\textit{AESC}). These tasks can be regarded as a token-level sequence labeling problem, and are generally tackled using supervised learning. The 2014 and 2015 SemEval workshops, co-located with COLING 2014 and NAACL 2015 respectively, included shared tasks on ABSA  \cite{pontiki_semeval_2014} and also followed this approach, which has also served as a way to encourage developments alongside this line of research \cite{mitchell_open_2013,irsoy_opinion_2014,liu_fine_grained_2015,zhang_neural_2015}.

The flexibility provided by the deep learning setting has helped multi-modal approaches to bloom. Examples of this include tasks such as machine translation \cite{specia_shared_2016,elliott_findings_2017}, word sense disambiguation \cite{chen_sense_2015}, visual question answering \cite{chen_reading_2017}, language grounding \cite{beinborn_multimodal_nodate,lazaridou_combining_2015}, and sentiment analysis \cite{poria_deep_2015,zadeh_mosi:_2016}. Specifically in this last example, the task focuses on generalizing text-based sentiment analysis to opinionated videos, where three communicative modalities are present: language (spoken words), visual (gestures), and acoustic (voice).

Although reviews often come under the form of a written commentary, people are increasingly turning to video platforms such as YouTube looking for product reviews to help them shop. In this context, \citet{marrese-taylor_mining_2017} explored a new direction, arguing that video reviews are the natural evolution of written product reviews and introduced a dataset of annotated video product review transcripts. Similarly, \citet{garcia_multimodal_2019} recently presented an improved version of the POM movie review dataset \cite{park_computational_2014}, with annotated fine-grained opinions.


Although the videos in these kinds of datasets represent a rich multi-modal source of opinions, the features of the language in them may fundamentally differ from written reviews given that information is conveyed through multiple channels (one for speech, one for gestures, one for facial expressions, one for vocal inflections, etc.) In these, different information channels complement each other to maximize the coherence and clarity of their message. This means that although the content of each channel may be comprehended in isolation, in theory we need to process the information in all the channels simultaneously to fully comprehend the message \cite{hasan-etal-2019-ur}. In this context, information extracted from nonverbal language in videos, such as gestures and facial expressions, as well as from audio in the manner of voice inflections or pauses, and from scenes, object or images in the video, become critical for performing well.


In light of this, our paper introduces a multi-modal approach for fine-grained opinion mining. We conduct extensive experiments on two datasets built upon transcriptions of video reviews, Youtubean \cite{marrese-taylor_mining_2017} and a fine-grain annotated version of the Persuasive Opinion Multimedia (POM) dataset \cite{park_computational_2014,garcia_multimodal_2019}, adapting them to our setting by associating timestamps to each annotated sentence using the video subtitles. Our results demonstrate the effectiveness of our proposed approach and show that by leveraging the additional modalities we can consistently obtain better performance.

\section{Related Work}
Our work is related to aspect extraction using deep learning, a task that is often tackled as a sequence labeling problem. In particular, our work is related to \citet{irsoy_opinion_2014}, who pioneered in the field by using multi-layered RNNs. Later, \citet{liu_fine_grained_2015} successfully adapted the architectures by \citet{mesnil_investigation_2013} which were originally developed for slot-filling in the context of Natural Language Understanding.

Literature offers related work on the usage of RNNs for open domain targeted sentiment \cite{mitchell_open_2013}, where \citet{zhang_neural_2015} experimented with neural CRF models using various RNN architectures on a dataset of informal language from Twitter.

Regarding target-based sentiment analysis, the literature contains several ad-hoc models that account for the sentence structure and the position of the aspect on it \cite{tang_effective_2016,tang_aspect_2016}. These approaches mainly use attention-augmented RNNs for solving the task. However, they require the location of the aspect to be known in advance and therefore are only useful in pipeline models, while instead we model aspect extraction and sentiment classification as a joint task or using multi-tasking.

\textit{AESC} has also often been tackled as a sequence labeling problem, mainly using Conditional Random Fields (CRFs) \cite{mitchell_open_2013}. To model the problem in this fashion, collapsed or sentiment-bearing IOB labels \cite{zhang_neural_2015} are used. Pipeline models (i.e. task-independent model ensembles) have also been extensively studied by the same authors. \citet{xu_joint_2014} performed \textit{AESC} by modeling the linking relation between aspects and the sentiment-bearing phrases.

When it comes to the video review domain, there is related work on YouTube mining, mainly focused on exploiting user comments. For example, \citet{wu2014crowd} exploited crowdsourced textual data from time-synced commented videos, proposing a temporal topic model based on LDA. \citet{tahara2010nicoscene} introduced a similar approach for \textit{Nico Nico}, using time-indexed social annotations to search for desirable scenes inside videos.

On the other hand, \citet{severyn_opinion_2014} proposed a systematic approach to mine user comments that relies on tree kernel models. Additionally, \citet{krishna2013polarity} performed sentiment analysis on YouTube comments related to popular topics using machine learning techniques, showing that the trends in users' sentiments is well correlated to the corresponding real-world events. \citet{siersdorfer2010howuseful} presented an analysis of dependencies between comments and comment ratings, proving that community feedback in combination with term features in comments can be used for automatically determining the community acceptance of comments.

We also find some papers that have successfully attempted to use closed caption mining for video activity recognition \cite{gupta2010usingclosed} and scene segmentation \cite{gupta2009usingclosed}. Similar work has been done using closed captions to classify movies by genre \cite{brezale2007using} and summarize video programs \cite{brezale2007using}.
Regarding multi-modal approaches for sentiment analysis, we see that previous work has focused mainly on sentiment classification, or the related task of emotion detection \cite{lakomkin_2017}, where the CMU MOSI dataset \cite{zadeh_mosi:_2016} appears as the main resource. In this setting, the main problem is how to model and capture cross-modality interactions to predict the sentiment correctly. In this regard \citet{zadeh_tensor_2017} proposed a tensor fusion layer that can better capture cross-modality interactions between text, audio and video inputs, while \citet{poria_context-dependent_2017} modeled inter-dependencies across difference utterances of a single video, obtaining further improvements.

\citet{blanchard_getting_2018} are, to the best of our knowledge, the first to tackle scalable multi-modal sentiment classification using both visual and acoustic modalities. More recently \citet{ghosal_contextual_2018} proposed an RNN-based multi-modal approach that relies on attention to learn the contributing features among multi-utterance representations. On the other hand \citet{pham_seq2seq2sentiment:_2018} introduced multi-modal sequence-to-sequence models which perform specially well in bi-modal settings. Finally,  \citet{akhtar_multi-task_2019} proposed a multi-modal, multi-task approach in which the inputs from a video (text, acoustic and visual frames), are exploited for simultaneously predicting the sentiment and expressed emotions of an utterance. Our work is related to all of these approaches, but it is different in that we apply multi-modal techniques not only for sentiment classification, but also for aspect extraction.

\begin{figure*}
    \centering
    \includegraphics[width=0.7\textwidth]{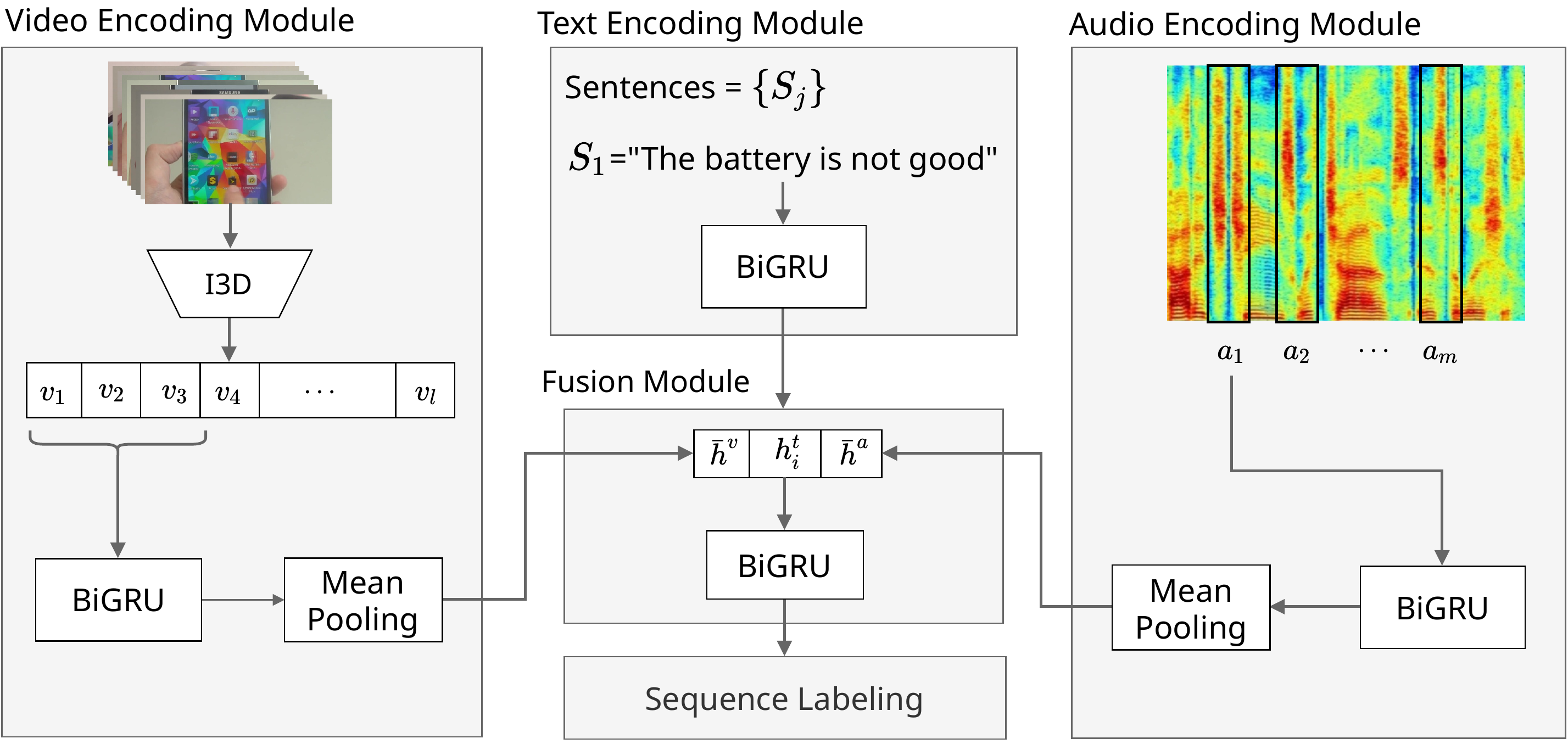}
    \caption{Overview of our proposed approach for multi-modal opinion mining}
    \label{fig:model}
\end{figure*}

Finally, \citet{marrese-taylor_mining_2017} and \citet{garcia_multimodal_2019} contributed multi-modal datasets obtained from product and movie reviews respectively, specifically for the task of fine-grained opinion mining. Furthermore, \citet{garcia_token_2019} recently used the latter to propose a hierarchical multi-modal model for opinion mining. Compared to them, our approach follows a more traditional setting for fine-grained opinion mining, while also offering a more general framework for the problem. \citet{garcia_token_2019} utilize a single encoder that receives as input the concatenation of the features for each modality, for each token. This requires explicit alignment between the features of the different modalities at the token level. In contrast, since each modality is encoded separately in our approach, we only require the feature alignment to be at the sentence level.

\section{Task Description}

Opinion mining can be performed at several levels of granularity, the most common ones being the sentence level, and the more fine-grained aspect level. Fine-grained opinion mining can be further subdivided in two tasks: aspect extraction and aspect-level sentiment classification. The former deals with finding the aspects being referred to, and the latter with associating them with a sentiment.

Previous work usually casts this task as a sequence-labeling problem, where models have to predict whether a token is a part of an aspect and infer its sentiment polarity \citep{mitchell_open_2013,zhang_neural_2015,liu_fine_grained_2015}. Depending on the dataset annotations, aspect categories are in some cases specified as well.

Formally, given a sentence $s = [ x_1, \ldots, x_n ]$, we want to automatically annotate each token $x_i$ with its aspect membership and polarity. In the simpler case where we only want to perform Aspect Extraction, a common annotation scheme is to tag each token with a label $y_i \in \sL^{\text{AE}}$ where $\sL^{\text{AE}}=\{I, O, B\}$. In this scheme, commonly known as IOB, $O$ labels indicate that a token is not a member of an aspect, $B$ labels indicate that a token is at the beginning of an aspect, and $I$ labels indicate that the token is inside an aspect.

Similarly, performing token-level Sentiment Classification only is equivalent to tagging each token with a label $y_i \in \sL^{\text{SC}}$ where $\sL^{\text{SC}}= \{\phi, +, -\}$, and $\phi$ denotes no sentiment, $+$ denotes a positive polarity and $-$ a negative one.

It is also possible to define a \textit{collapsed} annotation scheme, where aspect membership and sentiment polarity are encoded in a single tag. We define the label set for this setting as $\sL^{\text{C}} = \{O, B+, B-, I+, I-\}$.

\Cref{table:tags} shows the possible ways to annotate the sentence ``I love the saturated colors!'' under these three annotation schemes, where the aspect being referred to is ``saturated colors''.


\begin{table}[h!]
    \centering
    \scriptsize
    \begin{tabular}{c c c c c c c}
        \toprule
                          & \textbf{I} & \textbf{love} & \textbf{the} & \textbf{saturated} & \textbf{colors} & \textbf{!} \\
        \midrule
        $\sL^{\text{AE}}$ & $O$        & $O$           & $O$          & $B$                & $I$             & $O$        \\
        $\sL^{\text{SC}}$ & $\phi$     & $\phi$        & $\phi$       & $+$                & $+$             & $\phi$     \\
        $\sL^{\text{C}}$  & $O$        & $O$           & $O$          & $B+$               & $I+$            & $O$        \\
        \bottomrule
    \end{tabular}
    \caption{Label definition alternatives for the tasks in ABSA using sequence labeling.}%
    \label{table:tags}
\end{table}

Labels can be further augmented with type information. For example \citet{liu_fine_grained_2015} used different tags for opinion targets (e.g. B-TARG), and opinion expressions (e.g., B-EXPR), however, we do not rely on this information.

\section{Proposed Approach}

We propose a multi-modal approach for aspect extraction and sentiment classification that leverages video, audio and textual features. This approach assumes we have a video review $v$ containing opinions, its extracted audio stream $a$, and a transcription of the audio into a sequence of sentences $\sS$. Further, each sentence $s \in \sS$ is annotated with its respective start and end times in the video effectively mapping them to a video segment $v^s \subset v$ and its corresponding audio segment $a^s \subset a$. These segments do not necessarily cover the whole video i.e. $\cup \,v^s \subset v$ since the reviews may include parts that have no speech and therefore no sentences are associated to those. Our end goal is to produce a sequence of labels $l = [ y_1, \ldots, y_n]$ for each sentence $s = [x_1, \ldots, x_n]$ while exploiting the information contained in $v^s$ and $a^s$.




Figure \ref{fig:model} presents a high-level overview of our approach. We rely on an encoder-decoder paradigm to create separate representations for each modality \citep{cho_learning_2014}. The text encoding module generates a representation for each token in the input text, while the video and audio encoding layers produce utterance-level representations from each modality.

We propose combining these representations with an approach inspired by early-fusion \cite{xu_multilevel_2018}, which allows for the word-level representations to interact with audio and visual features. Finally, a sequence labeling module is in charge of taking the final token-level representations and producing a token-level label. In the following sub-sections we describe each component of our model.

\subsection{Text Encoding Module}

This module generates a representation of the natural language input so that the obtained representation is useful for the sequence labeling task. Our text encoder first maps each word $x_i$ into an embedded input sequence $\vx =  [\vx_1, \ldots, \vx_n ] $, then projects this into a vector $\vh_i^t \in \sR^{d_t}$, where $d_t$ corresponds to the hidden dimension of the obtained text representation. Although our text encoding module is generic, in this paper we implement it as a bi-directional GRU \citep{cho_learning_2014}, on top of pre-trained word embeddings, specifically GloVe \cite{pennington_2014_EMNLP_glove}, as follows.
\begin{equation}
    \vh_i^t = \text{BiGRU}(\vx_i, \vh^t_{i-1})
\end{equation}

\subsection{Audio Encoding Module}

We assume the existence of a finite set of time-ordered audio features $\va = [ \va_1, \ldots, \va_m ]$ extracted from each audio utterance $a^s$, for instance with the procedure described in Section \ref{subsec:implementation_details}. We feed these vectors into another bi-directional GRU to add context to each time step, obtaining hidden states $\vh^a_j \in \sR^{d_a}$.
\begin{equation}
    \vh_j^a = \text{BiGRU}(\va_j, \vh_{j-1}^a)
\end{equation}
To obtain  a condensed representation from the audio signal we again utilize mean pooling over the intermediate memory vectors, obtaining $\bar \vh^a$.

\subsection{Video Encoding Module}

We propose a video encoding layer that generates a visual representation summarizing spatio-temporal patterns directly from the raw input frames. Concretely, given a video segment $\vv = [ \vv_1, \ldots, \vv_T ]$, where $\vv_i$ is a vector representing a single frame in $v^s$, our encoding module first maps this sequence into another sequence of video features $\hat \vv = [ \hat \vv_1, \ldots, \hat \vv_l ]$ following the method described in Section \ref{subsec:implementation_details}. Later, this new sequence is mapped into a vector $\bar \vh^v \in \sR^{d_v}$ that captures summarized high-level visual semantics in the video, as follows:
\begin{equation}
    \vh_k^v = \text{BiGRU}(\hat \vv_k, \vh_{k-1}^v)
\end{equation}

\subsection{Fusion Module}

We utilize an early fusion strategy similar to \citet{xu_multilevel_2018} to aggregate the representations obtained from each modality. We concatenate the contextualized representation $\vh_i^t$ for each token to the summarized representations of the additional modalities, $\bar \vh^a$ and $\bar \vh^v$, and feed this final vector representation to an additional BiGRU:
\begin{equation}
    \vh_i = \text{BiGRU}([\vh_i^t; \bar \vh^a ; \bar \vh^v], \vh_{i-1})
\end{equation}
As a result, our model now allows the representation of each word in the input sentence to interact with the audio and visual features, enabling it to learn potentially different ways to associate each word with the additional modalities. An alternative way to achieve this would be to utilize attention mechanisms to enforce such association behavior, however, we instead let the model learn this relation without using any additional inductive bias.

\subsection{Sequence Labeling Module}

The main labeling module is a multi-layer perceptron guided by a self attention component. The self attention component enriches the representation $\vh_i$ with contextual information coming from every other sequence element by performing the following operations:
\begin{align}
    u_{i,j}      & = \vv_\alpha^{\top} \tanh(\mW_{\alpha} [\vh_i; \vh_j] + \vb_{\alpha}) \\
    \alpha_{i,j} & = \text{softmax}(u_{i,j})                                             \\
    \vt_i        & = \sum_{j=1}^n \alpha_{i,j} \cdot \vh_j                               \\
    \vo_i        & = \mW_l [\vh_i; \vt_i] + \vb_{l}
\end{align}
Where $\vo_i$ is a vector associated to input $x_i$, and $\vv_{\alpha}$, $\mW_{\alpha}$, $\mW_{l}$ and $\vb_{\alpha}$, $\vb_{l}$ are trainable parameters. As shown, these vectors are obtained using both the corresponding \textit{aligned} input $\vh_i$ and the attention-weighted vector $\vt_i$.

Following previous work, we feed these vectors into a Linear Chain CRF layer, which performs the final labeling. Neural CRFs have proven to be especially effective for various sequence segmentation or labeling tasks in NLP \cite{ma_end--end_2016,yang_ncrf++_2018,yang_design_2018-1}, and have also been used successfully in the past for open domain opinion mining \citep{zhang_neural_2015}. Concretely, we model emission and transition potentials as follows.
\begin{align}
    \psi_i     & := e(x_i, y_i;\theta) = \vh_i \cdot \vy_i \\
    \psi_{i,j} & := q(y_i, y_j;\mPi) = \mPi_{y_i, y_j}
\end{align}
Where $\vh_i$ is the fused hidden state for position i and $\theta$ denotes the parameters involved in computing this vector, $\vy_i$ is a one-hot vector associated to $y_i$, and $\mPi$ is a trainable matrix of size $\sL^{AE}$ or $\sL^{C}$ depending on the setting ---see Section \ref{sec:settings} for more details on this. The score function of a given input sentence $s$ and output sequence of labels $l$ is defined as:
\begin{equation}
    \Phi(s, l) = \sum \limits_{i=1}^n \log{e(x, y_i;\theta)} + \log{q(y_i,y_{i-1};\mPi)}
\end{equation}
In this work we directly optimize the negative log-likelihood associated to this score during training, and apply Viterbi decoding during inference to obtain the most likely labels.





\section{Experimental Setup}
\label{sec:settings}



We evaluate our proposal in several experimental settings based on previous work.

\begin{itemize}
    \item \textbf{Simple}: We only focus on the task of aspect extraction, following a sequence labeling approach with regular IOB tags in $\sL^{\text{AE}}$.

    \item \textbf{Collapsed Aspect-Level (CAL)}: We perform aspect extraction and aspect-level sentiment classification with a sequence labeling model, utilizing sentiment-bearing IOB tags in $\sL^{\text{C}}$.

    \item \textbf{Collapsed Sentence-Level (CSL)}: Like the previous setting, but we only keep sentence examples that contain a single sentiment, so we can perform sentence-level sentiment classification. Again, we use sequence labeling with sentiment-bearing IOB tags in $\sL^{\text{C}}$.

    \item \textbf{Joint Sentence-Level (JSL)}: We use a multi-tasking approach and perform sequence labeling for aspect extraction with regular IOB tags in $\sL^{\text{AE}}$, and sequence classification to predict the sentence-level sentiment. In this sense, we add a final 3-layer fully-connected neural network that receives a mean-pooled representation of the fusion layer $\bar{\vh} = \frac{1}{n}\sum_{i=1}^{n}{\vh_i}$ and predicts a sentence-level sentiment. As loss function we utilize the mini-batch average cross-entropy with the gold standard class label. The total loss is the sum of the losses for sequence labeling and sequence classification.
\end{itemize}

Previous work has also shown that most sentences present a single aspect, and therefore a single sentiment \citep{marrese-taylor_mining_2017,zuo_complementary_2015,zhao-EtAl:2010:EMNLP}, which motivates the introduction of the CSL and JSL settings. For these cases we filtered out sentences that do not fit this description.


\subsection{Data}
We report results on two different datasets containing fine-grained annotations for both opinion targets and sentiment.

First, we work with the Youtubean dataset \cite{marrese-taylor_mining_2017}, which contains sentences extracted from YouTube video annotated with aspects and their respective sentiments. The data comes from the user-provided closed-captions derived from 7 different long product review videos about a cell phone, totaling up to 71 minutes of audiovisual data. In total there are 578 long sentences from free spoken descriptions of the product, on average each sentence consist of 20 words. The dataset has a total of 525 aspects, with more than 66\% of the sentences containing at least one mention.

Second, we work with the fine-grained annotations gathered for the POM dataset by \citet{garcia_multimodal_2019}. This dataset is composed of 1000 videos containing reviews where a single speaker in frontal view makes a critique of a movie that he/she has watched. There are videos from 372 unique speakers, with 600 different movie titles being reviewed. Each video has an average length of about 94 seconds and contains 15.1 sentences on average. The fine-grained annotations we utilize are available for each token indicating if it is responsible for the understanding of the polarity of the sentence, and whether it describes the target of an opinion; each sentence has an average of 22.5 tokens. We assume that whenever there is an overlap between the span annotations for a given target and a certain polarity, the corresponding polarity can be assigned to that target, otherwise it is labeled as neutral.

Since the annotated sentences in both datasets are not associated to specific timestamps, in this work we propose a method based on heuristics to rescue the video segments that correspond to each annotated sentence by leveraging video subtitles (or closed-captions.)

\begin{figure}[h!]
    \scriptsize
    \centering
    \begin{lstlisting}
        168
        00:20:41,150 --> 00:20:45,109
        - How did he do that?
        - Made him an offer he could not refuse.
    \end{lstlisting}
    \caption{Excerpt of a subtitle chunk (in SubRip format,) showing its main components.}
    \label{fig:srt_example}
\end{figure}

As shown in Figure \ref{fig:srt_example}, closed captions or subtitles are composed of chunks that contain: (1) A numeric counter identifying each chunk, (2) The time at which the subtitle should appear on the screen followed by \texttt{-->} and the time when it should disappear, (3) The subtitle text itself on one or more lines, and (4) A blank line containing no text, indicating the end of this subtitle. These chunks exhibit a large variance in terms of their length, meaning that sentences are usually split into many chunks.

Starting from a subtitle file associated to a given product review video, we apply a fuzzy-matching approach between each annotated sentence for that review and each closed caption chunk. This is repeated for each one of the videos in our datasets. Whenever an annotated sentence matches exactly or has over 90\% similarity with a closed caption chunk, its time-span is associated to that sentence. Finally, the ``start'' and ``end'' timestamps assigned to each sentence are defined by the start and end time spans of their first and last associated closed captions, sorted by time.




\subsection{Implementation Details}
\label{subsec:implementation_details}

Pre-processing for the natural language input is performed utilizing spacy\footnote{\url{https://spacy.io}}, which we use mainly to tokenize. Input sentences are trimmed to a maximum length of 300 tokens, and tokens with frequency lower than 1 are replaced with a special \textit{UNK} marker. To work with the POM dataset, which is already tokenized, we first convert it to the ABSA format, which is tokenization agnostic, and then we process it.

Although our audio encoder is generic, in this work we follow \citet{lakomkin_2017} and use Fast Fourier Transform spectrograms to extract rich vectors from each audio segment. Specifically, we use a window length of 1024 points and 512 points overlap, giving us vectors of size 513. Alternative audio feature extractors such as \citet{degottex2014covarep} could also be utilized.

On the other hand, in this work we model video feature extraction using I3D \cite{carreira2017quo}. This method inflates the 2D filters of a well-known network e.g. Inception \cite{szegedy2015going, ioffe2015batch} or ResNet \cite{He_2016_CVPR} for image classification to obtain 3D filters, helping us better exploit the spatio-temporal nature of video. We first pre-process the videos by extracting features of size $1024$ using I3D with average pooling, taking as input the raw frames of dimension $256 \times 256$, at $25$ fps. We use the model pre-trained on the kinetics400 dataset \cite{Kinetics400} released by the same authors. Despite our choice to obtain video features, again we note that our video encoder is generic, so other alternatives such as C3D \cite{tran2015learning} could be utilized.

Finally, all of our models are trained in an end-to-end fashion using Adam \cite{kingma_adam} with a learning rate of $10^{-3}$. To prevent over-fitting, we add dropout to the text encoding layer. We use a batch size of 8 for the Youtubean dataset, and of 64 for the POM dataset. The language encoder uses a hidden state of size $150$, and we fine-tune the pre-trained GloVe.

On each case we compare the performance of our proposed approach against a baseline model that does not consider multi-modality, does not utilize pre-trained GloVe word embeddings and is based on a cross-entropy loss, in which case we simply utilize the mini-batch average cross-entropy between $\hat \vy_i = \text{softmax}(\vo_i)$ and the gold standard one-hot encoded labels $\vy_i$, a vector that is the size of the tag label vocabulary for the corresponding task.

\begin{table*}[t]
    \footnotesize
    \centering
    \begin{tabular}{cccccccc}
        \toprule
        \multirow{2}{*}{\textbf{Setting}} & \multirow{2}{*}{\textbf{Model}} & \multicolumn{3}{c}{\textbf{Aspect Extraction}} & \multicolumn{3}{c}{\textbf{Sentiment Classification}}                                                                         \\
        \cmidrule(lr){3-5}
        \cmidrule(lr){6-8}
                                          &                                 & \textbf{P}                                     & \textbf{R}                                            & \textbf{F1}        & \textbf{P}     & \textbf{R}     & \textbf{F1}    \\
        \midrule
        \multirow{2}{*}{Simple}
                                          & Baseline                        & 0.531                                          & 0.542                                                 & 0.533              & -              & -              & -              \\
                                          & Ours                            & \textbf{0.602**}                               & \textbf{0.568}                                        & \textbf{ 0.584***} & -              & -              & -              \\
        \midrule
        \multirow{2}{*}{CAL}
                                          & Baseline                        & 0.546                                          & 0.538                                                 & 0.539              & 0.710          & 0.688          & 0.696          \\
                                          & Ours                            & \textbf{0.590}                                 & \textbf{0.572}                                        & \textbf{0.581*}    & \textbf{0.722} & \textbf{0.722} & \textbf{0.718} \\
        \midrule
        \multirow{2}{*}{CSL}
                                          & Baseline                        & 0.526                                          & 0.463                                                 & 0.490              & \textbf{0.746} & \textbf{0.722} & \textbf{0.724} \\
                                          & Ours                            & \textbf{0.563}                                 & \textbf{0.581***}                                     & \textbf{0.568**}   & 0.720          & 0.674          & 0.688          \\
        \midrule
        \multirow{2}{*}{JSL}
                                          & Baseline                        & 0.483                                          & 0.521                                                 & 0.496              & \textbf{0.946} & \textbf{0.946} & \textbf{0.946} \\
                                          & Ours                            & \textbf{0.544***}                              & \textbf{0.552}                                        & \textbf{0.545***}  & \textbf{0.946} & \textbf{0.946} & \textbf{0.946} \\
        \bottomrule
    \end{tabular}
    \caption{Summary of our results on the Youtubean dataset, *** denotes statistical significance at 99\% confidence, ** at 95\% and * at 90\%.}
    \label{table:results_youtubean}
\end{table*}

\begin{table*}[t]
    \centering
    \footnotesize
    \begin{tabular}{cccccccc}
        \toprule
        \multirow{2}{*}{\textbf{Setting}} & \multirow{2}{*}{\textbf{Model}} & \multicolumn{3}{c}{\textbf{Aspect Extraction}} & \multicolumn{3}{c}{\textbf{Sentiment Classification}}                                                                                \\
        \cmidrule(lr){3-5}
        \cmidrule(lr){6-8}
                                          &                                 & \textbf{P}                                     & \textbf{R}                                            & \textbf{F1}      & \textbf{P}        & \textbf{R}        & \textbf{F1}       \\
        \midrule
        \multirow{2}{*}{Simple}
                                          & Baseline                        & 0.394                                          & 0.379                                                 & 0.386            & -                 & -                 & -                 \\
                                          & Ours                            & \textbf{0.396}                                 & \textbf{0.406}                                        & \textbf{0.399}   & -                 & -                 & -                 \\
        \midrule
        \multirow{2}{*}{CAL}
                                          & Baseline                        & 0.364                                          & \textbf{0.401*}                                       & 0.382            & \textbf{0.540***} & 0.416             & 0.270             \\
                                          & Ours                            & \textbf{0.444**}                               & 0.368                                                 & \textbf{0.402**} & 0.488             & \textbf{0.466***} & \textbf{0.342***} \\
        \midrule
        \multirow{2}{*}{CSL}
                                          & Baseline                        & 0.387                                          & 0.375                                                 & \textbf{0.408*}  & \textbf{0.614}    & \textbf{0.446}    & 0.296             \\
                                          & Ours                            & \textbf{0.438*}                                & \textbf{0.378}                                        & 0.404            & 0.532             & \textbf{0.446}    & \textbf{0.304}    \\
        \midrule
        \multirow{2}{*}{JSL}
                                          & Baseline                        & 0.381                                          & 0.357                                                 & 0.367            & 0.798             & 0.802             & 0.788             \\
                                          & Ours                            & \textbf{0.442***}                              & \textbf{0.401*}                                       & \textbf{0.420*}  & \textbf{0.924***} & \textbf{0.924***} & \textbf{0.922***} \\
        \bottomrule
    \end{tabular}
    \caption{Summary of our results for the test set of the POM dataset, *** denotes statistical significance at 99\% confidence, ** at 95\% and * at 90\%.}
    \label{table:results_pom}
\end{table*}

\begin{table}[t]
    \footnotesize
    \centering
    \begin{tabular}{rcccc}
        \toprule
        \multirow{2}{*}{\textbf{Model}} & \multicolumn{3}{c}{\textbf{Aspect Extraction}}                            \\
        \cmidrule{2-4}
                                        & \textbf{P}                                     & \textbf{R} & \textbf{F1} \\
        \midrule
        T                               & 0.532                                          & 0.543      & 0.533       \\
        T + CRF                         & 0.558                                          & 0.528      & 0.541       \\
        T + GV                          & 0.562                                          & 0.537      & 0.548       \\
        T + GV + CRF                    & 0.576*                                         & 0.569      & 0.571**     \\
        T + A + V                       & 0.587*                                         & 0.578      & 0.580*      \\
        T + CRF + A + V                 & 0.578                                          & 0.570      & 0.573*      \\
        T + GV + CRF + A + V            & 0.602**                                        & 0.568      & 0.584***    \\
        \bottomrule
    \end{tabular}
    \caption{Ablation study on aspect extraction on the simple setting. *** denotes differences against the only text model (T) results are statistically significant at 99\% confidence, ** at 95\% and * at 90\%. (A + V) refers to the audio and video modalities, (GV) stands for GLoVe embeddings and (CRF) for the model trained using the Conditional Random Fields loss.}
    \label{table:ablation}
\end{table}
\subsection{Evaluation}

Since the size of Youtubean is relatively small, all our experiments in this dataset are evaluated using 5-fold cross validation. In the case of the POM dataset, we report performance on the validation and test sets averaging results for 5 different random seeds. In both cases we compare models using paired two-sided t-tests to check for statistical significance of the differences.

To evaluate our sequence labeling tasks we used the CoNLL \textit{conlleval} script, taking the aspect extraction F1-score as our model selection metric for early stopping. To perform joint aspect extraction and sentiment classification, we considered \textit{positive}, \textit{negative} and \textit{neutral} as sentiment classes, and decoupled the IOB collapsed tags using simple heuristics. Concretely, we recover the aspect extraction F1-score as well as classification performances for each sentiment class.

\section{Results}

To evaluate the effectiveness of our proposals, we perform several ablation studies on the \textit{Simple} setting for the Youtubean dataset. Using variations of our baseline with pre-trained GLoVe embeddings (GV), conditional random field (CRF), audio and video modalities (A+V). Experiments are also performed using 5-fold cross-validation, and comparisons are always tested for significance using paired two-sided t-tests. 


As Table \ref{table:ablation} shows, although every proposed model variation performs better than the baseline, only the model uses video and audio modalities obtains a statistically superior performance. We also see that our proposed multi-modal variation is the one that obtains the best performance, also being statistically significant at the highest level of confidence. We believe these results show that our proposed multi-modal architecture is not only able to exploit the features in the audio and video inputs, but it can also leverage the information in the pre-trained word embeddings and benefit from having an inductive bias that is tailored for the task at hand, in this case, with a loss based on structured prediction for sequence labeling.

\begin{table}[t]
    \centering
    \footnotesize
    \begin{tabular}{cccc}
        \toprule
        \textbf{Setting} & \textbf{Model} & \textbf{AE F1}    & \textbf{SC F1}    \\
        \midrule
        \multirow{2}{*}{Simple}
                         & Baseline       & 0.428             & -                 \\
                         & Ours           & \textbf{0.433}    & -                 \\
        \midrule
        \multirow{2}{*}{CAL}
                         & Baseline       & 0.412             & 0.240             \\
                         & Ours           & \textbf{0.427}*** & \textbf{0.310}**  \\
        \midrule
        \multirow{2}{*}{CSL}
                         & Baseline       & 0.408             & \textbf{0.264}    \\
                         & Ours           & \textbf{0.423}*   & 0.262             \\
        \midrule
        \multirow{2}{*}{JSL}
                         & Baseline       & 0.387             & \textbf{0.950}*** \\
                         & Ours           & \textbf{0.469}**  & 0.840             \\
        \bottomrule
    \end{tabular}
    \caption{Results for the validation set of the POM dataset, where *** denotes results are statistically significant at 99\% confidence, ** at 95\% and * at 90\%.}
    \label{table:results_pom_valid}
\end{table}
Table \ref{table:results_youtubean} summarizes our results for the Youtubean dataset, where we can see that our proposed multi-modal approach is able to outperform the baseline model for all settings in the aspect extraction task. When it comes to sentiment classification, our multi-modal approaches do not obtain significant performance gain in all cases, sometimes performing worse although without statistical significance.
\begin{figure*}
    \centering
    \includegraphics[width=\textwidth]{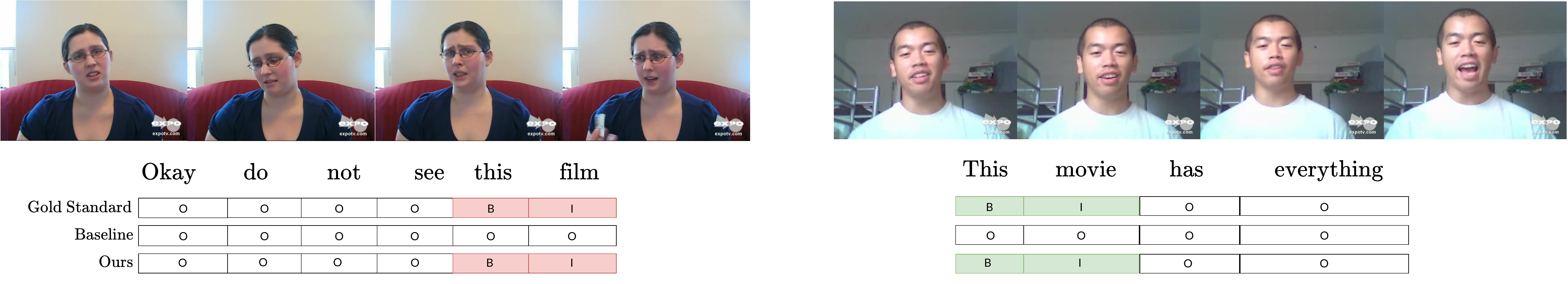}\\
    \caption{Qualitative comparison between baseline and our method on the POM dataset. Green and red boxes represent positive and negative sentiment respectively.}
    \label{fig:qualitative1}
\end{figure*}
\begin{figure*}
    \centering
    \includegraphics[width=\textwidth]{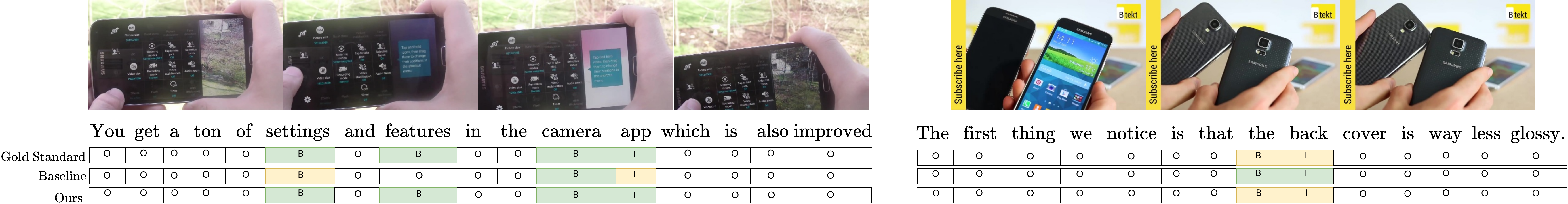}\\
    \caption{Qualitative comparison between baseline and our method on the Youtubean dataset. Green and yellow boxes represent positive and neutral sentiment respectively.}
    \label{fig:qualitative2}
\end{figure*}
We also compare our results to the performance reported by \citet{marrese-taylor_mining_2017}, who experimented on the \textit{Simple} and \textit{CSL} settings. Their models also use pre-trained word embedding ---although different from GloVe--- and as input they additionally receives binary features derived from POS tags and other word-level cues. We note, however, that they only experimented with a maximum length of 200 tokens, which makes our results not directly comparable. Their performance on aspect extraction for the \textit{Simple} and \textit{CAL} tasks are 0.561 and 0.555 F1-Score respectively, both of which are lower than ours. In terms of sentiment classification, they report results for each sentiment class with F1-Scores of 0.523, 0.149 and 0.811 for the positive, negative and neutral classes, respectively. Our model is able to outperform this baseline, with a cross-class average F1-Score of 0.718. We do not deepen the analysis in this regard, as numbers are difficult to interpret without statistical testing.

Table \ref{table:results_pom_valid} and Table \ref{table:results_pom} summarize our results for the \textit{POM} dataset for the validation and test splits respectively. Compared to the previous dataset we see similar results where our multi-modal approach consistently outperforms the baseline for aspect extraction, but with the gains being comparatively smaller. We also see that our model is able to significantly outperform the baseline in the sentiment classification tasks at least in two of out the three settings. In terms of previous work, our results cannot be directly compared to \citet{garcia_token_2019} and \citet{garcia_multimodal_2019} as their problem setting is different from ours.

On a more broad perspective, we think the performance differences across datasets are related to the nature of each dataset. Meanwhile Youtubean contains reviews about actual physical products, which are often shown in the videos at the same time the reviewer is speaking, the POM dataset contains movie reviews where the speakers directly face the camera during most of the video, without utilizing any additional support material. As a result, the video reviews in the Youtubean dataset mainly focus on capturing images of the products under discussion, with relatively fewer scenes showing the reviewer. This means that there may be few visual cues in the manner of facial expressions or other specific actions that the models could exploit in order to perform better at the sentiment classification task, but more cues useful for aspect extraction. This situation is reverted in the POM dataset, which could explain why our models tend to perform better for sentiment classification, but offering smaller gains for the AE task.

We also think performance differences across datasets are to some extent explained by the nature of the annotations on each case. The annotation guidelines utilized to elaborate each dataset are actually quite different, with the annotations in the Youtubean dataset closely following those of the well-known SemEval datasets, which are target-centric and the POM standards substantially diverging from this. Concretely, \citet{garcia_multimodal_2019} propose a two-level annotation method,  where ``the smallest span of words that contains all the words necessary for the recognition of an opinion'' are to be annotated. As a result, aspects annotated in the POM dataset often include pronouns which are more difficult to identify as aspects, often requiring co-reference resolution. With regards to aspect polarity, while it can be extracted directly from the Youtubean annotations, in the case of POM we needed  some pre-processing as target and sentiment are annotated using independent text spans.

Qualitative results of the POM and Youtubean dataset in a multitask CAL can be seen in Figure \ref{fig:qualitative1} and \ref{fig:qualitative2} respectively, results suggest that the method learn to use the information from additional modalities and enhance the sentiment and aspect prediction.

Finally, as we observe that our models tend to obtain bigger gains on the AE tasks rather than on SC, we think this behavior can be partially attributed to the inductive bias of our model, which makes it specially suitable for sequence segmentation tasks.

\section{Conclusions}

In this paper we have presented a multi-modal approach for fine-grained opinion mining, introducing a modular architecture that utilizes features derived from the audio, video frames and language transcription of video reviews to perform aspect extraction and sentiment classification at the sentence level. To test our proposals we have taken two datasets built upon video review transcriptions containing fine-grained opinions, and introduced a technique that leverages the video subtitles to associate timestamps to each annotated sentence. Our results offer empirical evidence showing that the additional modalities contain useful information that can be exploited by our models to offer increased performance for both aspect extraction and sentiment classification, consistently outperforming text-only baselines.

For future work, we are interested in exploring other ways to capture cross-modal interactions, exploit the temporal relationship between the representations of different modalities, and test alternative ways to better deal with our multi-task settings.

\section*{Acknowledgments}
We are grateful for the support provided by the NVIDIA Corporation, donating two of the GPUs used for this research.

\bibliography{bibliography}
\bibliographystyle{acl_natbib}

\end{document}

%% file: math_commands.tex
\def\va{{\bm{a}}}
\def\vb{{\bm{b}}}

\def\vh{{\bm{h}}}

\def\vo{{\bm{o}}}

\def\vt{{\bm{t}}}

\def\vv{{\bm{v}}}

\def\vx{{\bm{x}}}
\def\vy{{\bm{y}}}

\def\mW{{\bm{W}}}

\def\mPi{{\bm{\Pi}}}

\DeclareMathAlphabet{\mathsfit}{\encodingdefault}{\sfdefault}{m}{sl}
\SetMathAlphabet{\mathsfit}{bold}{\encodingdefault}{\sfdefault}{bx}{n}

\def\sL{{\mathbb{L}}}

\def\sR{{\mathbb{R}}}
\def\sS{{\mathbb{S}}}

